\documentclass{llncs}

\usepackage{tikz,multicol,caption}

\usetikzlibrary{backgrounds,shapes,arrows,positioning,calc,snakes,fit,automata}
\usepackage{epsfig}
\usepackage{multicol}
\usepackage{colortbl}
\usepackage{color,colortbl}
\usepackage{makecell}
\usepackage{enumitem}
\usepackage{pgfplots}

\usepackage{pslatex,amssymb,amsmath}

\usepackage{linguex}

\newcommand{\hidden}[1]{}

\newcommand{\res}{\mathsf{res}}

\newcommand{\rel}{\mathcal{R}}
\newcommand{\jury}{\mathcal{J}}
\newcommand{\game}{\mathcal{G}}

\newcommand{\type}{T}
\newcommand{\Type}{\mathcal{T}}
\newcommand{\tp}{t}
\newcommand{\tj}{{\sf tj}}

\newcommand{\belief}{\beta}
\newcommand{\bbelief}{\hat\beta}
\newcommand{\intpr}{\xi}
\newcommand{\event}{E}

\newcommand{\Strat}{S}

\newcommand{\ulf}{{\sc ulf}}

\newcommand{\turn}[1]{\langle#1\rangle}

\newcommand{\play}{\rho}
\newcommand{\vocab}{(V_0\cup V_1)}
\newcommand{\mevocab}{(V_0\cup V_1)^\omega}

\newcommand{\Play}{{\mathcal P}}

\newcommand{\strat}{\sigma}

\newcommand*{\myfont}{\fontfamily{pzc}\selectfont}

\newcommand{\bck}{{\sf bkgnd}}
\newcommand{\iqap}{{\sf iqap}}

\newcommand{\corr}{{\sf corr}}
\newcommand{\exptn}{{\sf exp}}
\newcommand{\confq}{{\sf confQ}}

\newcommand{\SP}[1]{\textcolor{blue}{S: #1}}

\newcommand{\hist}{\mathfrak{h}}
\newcommand{\win}{\mathit{Win}}
\def\refname{\large {\bf References}}
\begin{document}

\title{Bias in Semantic and Discourse Interpretation}
\author{Nicholas Asher\inst{1} \and Soumya Paul\inst{2}}
\authorrunning{Asher et al.}

\institute{
  IRIT, Universit\'e Paul Sabatier, Toulouse, France\\
  \email{nicholas.asher@irit.fr}
\and
SnT, Universit\'e du Luxembourg, Esch-sur-Alzette, Luxembourg\\
\email{soumya.paul@gmail.com}
}
\date{}

\maketitle

\vspace{-0.2cm}

\begin{abstract}
In this paper, we show how game-theoretic work on conversation combined with a theory of discourse structure provides
a framework for studying interpretive bias.  Interpretive bias is an essential feature of learning and understanding but also something that can be used to pervert or subvert the truth.  The framework we develop here provides tools for understanding and analyzing the range of interpretive biases and the factors that contribute to them.
\end{abstract}
  \section{Introduction}\label{sec:aims}


\textit{Bias} is generally considered to be a negative term: a biased story is seen as one that perverts or subverts the truth by offering a partial or incomplete perspective on the facts.  But bias is in fact essential to understanding: one cannot interpret a set of facts---something humans are disposed to try to do  even in the presence of data that is nothing but noise [38]---without relying on a bias or hypothesis to guide that interpretation.  

Suppose someone presents you with the sequence  $0, 1, 1, 2, 3$ and tells you to guess the next number. To make an educated guess, you must understand this sequence as instantiating a particular pattern; otherwise, every possible continuation of the sequence will be equally probable for you. Formulating a hypothesis about what pattern is at work will allow you to predict how the  sequence will play out, putting you in a position to make a reasonable guess as to what comes after $3$. Formulating the hypothesis that this sequence is structured by the Fibonacci function (even if you don't know its name), for example, will lead you to guess that the next number is 5; formulating the hypothesis that the sequence is structured by the successor function but that every odd successor is repeated once will lead you to guess that it is 3. Detecting a certain pattern allows you to determine what we will call a \textit{history}: a  set of given entities or eventualities and a set of relations linking those entities together.   The  sequence of numbers $0, 1, 1, 2, 3$ and the set of  relation instances that the Fibonacci sequence entails as holding between them is one example of a history.  \textit{Bias}, then, is the set of features, constraints, and assumptions that lead an interpreter to select one history---one way of stitching together a set of observed data---over another.\footnote{In machine learning and statistics, `bias' often refers to the divergence between an estimated hypothesis about a parameter and its objective value.  We use `bias' to refer to the features and assumptions that lead one to formulate a hypothesis in the first place.}

Bias is also operative in linguistic interpretation.  An interpreter's bias surfaces, for example, when the interpreter connects bits of information content together to resolve ambiguities. 
  Consider:
  \ex. \label{julie} Julie isn't coming.  The meeting has been cancelled.  
 
 \noindent
While these clauses are not explicitly connected, an  interpreter will typically have antecedent  biases that  lead her to interpret eventualities described by the two clauses as figuring in one of two histories: one in which the eventuality described by the first clause caused the second, or one in which the second caused the first.  
  Any time that structural connections are left implicit by speakers---and this is much if not most of the time in text--- interpreters will be left to infer these connections and thereby potentially create their own history or version of events. 
  

Every model of data, every history over that data, comes with a bias that allows us to use observed facts
to make predictions; bias even determines what kind of predictions the model is meant to make.
Bayesian inference, which underlies many powerful models of inference and machine learning, likewise relies on bias in several ways: the estimate of a state given evidence depends upon a prior probability distribution over states, on assumptions about what parameters are probabilistically independent, and on assumptions about the kind of conditional probability distribution that each parameter abides by (e.g., normal distribution, noisy-or, bimodal). Each of these generates a (potentially different) history.


\subsection{Objective of the paper}
In this paper, we propose a program for research on bias.  We will show how to model various types of bias as well as the way in which bias leads to the selection of a history for a set of data, where the data might be a set of nonlinguistic entities or a set of linguistically expressed contents.  In particular, we'll look at  what people call ``unbiased'' histories.  For us these also involve a bias, what we call a ``truth seeking bias''.  This is a bias that gets at the truth or acceptably close to it.  Our model can show us what such a bias looks like.  And we will examine the question of whether it is possible to find such a truth oriented bias for a set of facts, and if so, under what conditions.   Can we detect and avoid biases that don't get at the truth but are devised for some other purpose?

Our study of interpretive bias relies on three key premises.  The first premise is that histories are {\em discursive} interpretations of a set of data in the sense that like discourse interpretations, they link together a set of entities with semantically meaningful relations.  As such they are amenable to an analysis using the tools used to model a discourse's content and structure.  The second is that a bias consists of a \textit{purpose} or goal that the histories it generates are built to achieve and that agents build histories for many different purposes---to discover the truth or to understand, but also to conceal the truth, to praise or disparage, to persuade or to dissuade.  To properly model histories and the role of biases in creating them, we need a model of the discourse purposes to whose end histories are constructed and of the way that they, together with prior assumptions, shape and determine  histories.  
The third key premise of our approach is that bias is manifested in and conveyed through {\em histories}, and so studying histories is crucial for a better understanding of bias.

\subsection{Some examples of bias}
Let's consider the following example of biased interpretation of a conversation.
Here is an example analyzed in \cite{JOLLI} to which we will return in the course of the paper.
\ex.  \label{Sheehan}
\a. \label{Sa} {\bf Reporter:} On a different subject is there a reason that the Senator won't say whether or not someone else bought some suits for him?
\b. \label{Sb} {\bf Sheehan:} Rachel, the Senator has reported every gift he has ever received.
\c. \label{Sc} {\bf Reporter:} That wasn't my question, Cullen.
\d. \label{Sd} {\bf Sheehan:} (i) The Senator has reported every gift he has ever received. (ii)  We are not going to respond to unnamed sources on a blog.
\e. \label{Se}  {\bf Reporter:} So Senator Coleman's friend has not bought these suits for him? Is that correct?
\f.  \label{Sf} {\bf Sheehan:} The Senator has reported every gift he has ever received.

Sheehan continues to repeat, {\myfont ``The Senator has reported every gift he has ever received''} seven more times in two minutes to every follow up question by the reporter corps. {\tt http://www.youtube.com/watch?v=VySnpLoaUrI}. For convenience, we denote this sentence uttered by Sheehan (which is an EDU in the languare of SDRT as we shall see presently) as $\alpha$. 
\medskip

Now imagine two ``juries,'' onlookers or judges who interpret what was said and evaluate the exchange, yielding differing interpretations.  The interpretations differ principally in how the different contributions of Sheehan and the reporter hang together.  In other words, the different interpretations provide different discourse structures that we show schematically in the graphs below.  The first is one in which Sheehan's response $\alpha$ in \ref{Sheehan}b is somewhat puzzling and not taken as an answer to the reporter's question in \ref{Sheehan}a.  In effect this ``jury'' could be the reporter herself.  This Jury then interprets the move in \ref{Sheehan}c as a correction of the prior exchange. 
The repetition of $\alpha$ in \ref{Sheehan}d.ii is taken tentatively as a correction of the prior exchange (that is, the moves \ref{Sheehan}a, \ref{Sheehan}b and \ref{Sheehan}c together), which the Jury then takes the reporter to try to establish with \ref{Sheehan}e.   
\hidden{
\begin{center}
\begin{tikzpicture}[lbl/.style={font=\scriptsize\sf}]
    \node(q1) at (0,0) {{\bf a}};
    \node(ind) at (0,-2) {{\bf $\alpha$}};
    \node(corr) at (2,-1) {{\bf c}};
    \node(corr2) at (4,-1) {{\bf $\alpha$}};
    \node(exp) at (6,-1) {{\bf d(ii)}};
    \node(dir) at (8,-1) {{\bf e}};
    \draw[->] (q1) -- (ind) node[lbl,midway,fill=white]{bkgd};
    \node(cdu1) [draw=black!50,thick,fit={(q1)(ind)}]{};
    \draw[->] (cdu1) -- (corr) node[lbl,above,midway]{corr};
    \draw[->] (ind) to [bend right=15] node[lbl,below,midway]{reit} (corr2);
    \node(cdu2) [draw=black!50,thick,fit={(cdu1)(corr)}]{};
    \draw[->] (cdu2) -- (corr2) node[lbl,above,midway]{corr?};
    \node(cdu3) [draw=black!50,thick,fit={(cdu2)(corr2)}]{};
    \node(cdu4) [draw=black!50,thick,fit={(cdu3)(corr2)(exp)}]{};
    \draw[->] (cdu4) -- (dir) node[lbl,above,midway]{res};
    \draw[->] (cdu3) to node[lbl,above,midway]{exp} (exp);
\end{tikzpicture}
\end{center}
}
When Sheehan repeats \ref{Sheehan}a again in \ref{Sheehan}f, this jury might very well take Sheehan to be evading all questions on the subject.

A different Jury, however, might have a different take on the conversation as depicted in the discourse structure below.  Such a jury might take $\alpha$ to be at least an indirect answer to the question posed in \ref{Sheehan}a, and as a correction to the Reporter's evidently not taking $\alpha$ as an
answer.   The same interpretation of $\alpha$ would hold for this Jury when it is repeated in \ref{Sheehan}f.  Such a Jury would be a supporter of Sheehan or even Sheehan himself.
\hidden{
\begin{center}
\begin{tikzpicture}[lbl/.style={font=\scriptsize\sf}]
    \node(q1) at (0,0) {{\bf a}};
    \node(ind) at (0,-2) {{\bf $\alpha$}};
    \node(corr) at (2,-1) {{\bf c}};
    \node(corr2) at (4,-1) {{\bf $\alpha$}};
    \node(exp) at (6,-1) {{\bf d(ii)}};
    \node(dir) at (8,-1) {{\bf e}};
    \draw[->] (q1) -- (ind) node[lbl,midway,fill=white]{iqap};
    \node(cdu1) [draw=black!50,thick,fit={(q1)(ind)}]{};
    \draw[->] (cdu1) -- (corr) node[lbl,above,midway]{corr};
    \draw[->] (q1) to [bend left=15] node[lbl,above,midway]{iqap} (corr2);
    \node(cdu2) [draw=black!50,thick,fit={(cdu1)(corr)}]{};
    \draw[->] (cdu2) -- (corr2) node[lbl,above,midway]{corr};
    \node(cdu3) [draw=black!50,thick,fit={(cdu2)(corr2)}]{};
    \node(cdu4) [draw=black!50,thick,fit={(cdu3)(corr2)(exp)}]{};
    \draw[->] (cdu4) -- (dir) node[lbl,above,midway]{res};
    \draw[->] (cdu3) to node[lbl,above,midway]{exp} (exp);
\end{tikzpicture}
\end{center}
}
What accounts for these divergent discourse structures?  We will argue that it is the biases of the two Juries that create these different interpretations.  And these biases are revealed at least implicitly in how they interpret the story: Jury 1 is at the outset at least guarded, if not skeptical, in its appraisal of Sheehan's interest in answering the reporter's questions.  On the other hand, Jury 2 is fully convinced of Sheehan's position and thus interprets his responses much more charitably.  \cite{JOLLI} shows formally that there is a co-dependence between biases and interpretations; a certain interpretation created because of a certain bias can in turn strengthen that bias, and we will sketch some of the details of this story below.  

The situation of our two juries applies to a set of nonlinguistic facts.  In such a case we take our ``jury'' to be the author of a history over that set of facts.  The jury in this case evaluates and interprets the facts just as our juries did above concerning linguistic messages.   To tell a history about a set of facts is to connect them together just as discourse constituents are connected together.  And these connections affect and may even determine the way the facts are conceptualized \cite{hunter:etal:2017}. Facts typically do not wear their connections to other facts on their sleeves and so how one takes those connections to be is often subject to bias.  Even if their characterization and their connections to other facts are ``intuitively clear'', our jury may choose to pick only certain connections to convey a particular history or even to make up connections that might be different.  One jury might build a history over the set of facts that conveys one set of ideas, while the other might build a quite different history with a different message.  Such histories reflect the purposes and assumptions that were exploited to create that structure.  

As an example of this, consider the lead paragraphs of articles from the {\em New York Times}, {\em Townhall} and {\em Newsbusters} concerning the March for Science held in April, 2017.
\begin{quote}
\small{{\em The March for Science on April 22 may or may not accomplish the goals set out by its organizers. But it has required many people who work in a variety of scientific fields --- as well as Americans who are passionate about science --- to grapple with the proper role of science in our civic life.  The discussion was evident in thousands of responses submitted to NYTimes.com ahead of the march, both from those who will attend and those who are sitting it out.\vspace{-0.4cm}\begin{flushright}--New York Times\end{flushright}
      
  Do you have march fatigue yet? The left, apparently, does not, so we're in for some street theater on Earth Day, April 22, with the so-called March for Science.  It's hard to think of a better way to undermine the public's faith in science than to stage demonstrations in Washington, D.C., and around the country modeled on the Women's March on Washington that took place in January. The Women's March was an anti-Donald Trump festival.   
Science, however, to be respected, must be purely the search for truth. The organizers of this ``March for Science" -- by acknowledging that their demonstration is modeled on the Women's March -- are contributing to the politicization of science, exactly what true upholders of science should be at pains to avoid.\vspace{-0.4cm}\begin{flushright}--Townhall\end{flushright}  
      
Thousands of people have expressed interest in attending the “March for Science” this Earth Day, but internally the event was fraught with conflict and many actual scientists rejected the march and refused to participate.\vspace{-0.4cm}\begin{flushright}--Newsbusters\end{flushright}}}
\end{quote}
These different articles begin with some of the same basic facts: the date and purpose of the march, and the fact that the  march's import for the science community is controversial, for example.    But bias led the reporters to stitch together very different histories.  The \textit{New York Times}, for instance, interprets the controversy as generating a serious discussion about ``the proper role of science in our civic life,'' while \textit{Townhall} interprets the march as a political stunt that does nothing but undermine science. 

While the choice of wording helps to convey bias, just as crucial is the way that the reporters portray the march as being related to other events.  Which events authors choose to include in their history, which they leave out, and the way the events chosen relate to the march are crucial factors in conveying bias. 
{\em Townhall}'s 
bias against the March of Science expressed in the argument that it politicizes science cannot be traced back to negative opinion words; it relies on a comparison between the March for Science and the Women's March, which is portrayed as a political, anti-Trump event. 
 {\em Newsbusters} takes a different track: the opening paragraph conveys an overall negative perspective on the March for Science, despite its neutral language, but it achieves this by contrasting general interest in the march with a claimed negative view of the march by many ``actual scientists.'' On the other hand, the {\em New York Times} points to an important and presumably positive outcome of the march, despite its controversiality: a renewed look into the role of science in public life and politics. Like \textit{Newsbusters}, it lacks any explicit evaluative language and relies on the structural relations between events to convey an overall positive perspective; it contrasts the controversy surrounding the march with a claim that the march has triggered an important discussion, which is in turn buttressed by the reporter's mentioning of  the responses of the \textit{Times}' readership.

A formally precise account of interpretive bias will thus require an analysis of histories and their structure and to this end, we exploit Segmented Discourse Representation Theory or SDRT \cite{asher:1993,asher:lascarides:2003}. As the most precise and well-studied formal model of discourse structure and interpretation to date, SDRT enables us to characterize and to compare histories in terms of their structure and content.  But neither SDRT nor any other, extant theoretical or computational approach to discourse interpretation can adequately deal with the inherent subjectivity and interest relativity of interpretation, which our study of bias will illuminate.  {\em Message Exchange (ME) Games}, a theory of games that builds on SDRT, supplements SDRT with an analysis of the purposes and assumptions that figure in bias.  While epistemic game theory in principle can supply an analysis of these assumptions, it lacks linguistic constraints and fails to reflect the basic structure of conversations \cite{JPL}.     ME games will enable us not only to model the purposes and assumptions behind histories but also to evaluate their complexity and feasibility in terms of the existence of winning strategies.\footnote{The mathematical structure of ME games also makes it natural to investigate how ME game analyses of bias interact with information-theoretic analyses proposed by \cite{hilbert:2012} as a unifying factor of cognitive biases studied in psychology.}

Bias has been studied in cognitive psychology and empirical economics \cite{tversky:kahneman:1973,tversky:kahneman:1985,tversky:kahneman:1975,tversky:kahneman:1983,erev:etal:1994,hintzman:1984,hintzman:1988,hilbert:2012,wilkinson:klaes:2012}.  Since the seminal work of Kahneman and Tversky and the economist Allais, psychologists and empirical economists have provided valuable insights into cognitive biases in simple decision problems and simple mathematical tasks \cite{baron:2000}.  
Some of this work, for example the bias of framing effects \cite{tversky:kahneman:1985}, is directly relevant to our theory of interpretive bias.  A situation is presented using certain lexical choices that lead to different ``frames'': $x \%$ of the people will live if you do $z$ (frame 1)  versus $ y\%$ of the people will die if you do $z$ (frame 2).  In fact, $x + y = 100$, the total population in question; so the two consequents of the conditionals are equivalent.  Each frame elaborates or ``colors'' $z$ in a way that affects an interpreter's evaluation of $z$.  These frames are in effect short histories whose discourse structure explains their coloring effect.  
Psychologists, empirical economists and statisticians have also investigated cases of cognitive bias in which subjects {\em deviate} from prescriptively rational or independently given objective outcomes in quantitative decision making and frequency estimation, even though they arguably have the goal of seeking an optimal or ``true'' solution.  
In a general analysis of interpretive bias like ours, however, it is an open question whether there is an objective norm or not, whether it is attainable and, if so, under what conditions, and whether an agent builds a history for attaining that norm or for some other purpose. 

\subsection{Organization of the paper}
Our paper is organized as follows.  Section \ref{sec:model} introduces our model of interpretive bias.  Section \ref{sec:consequences} looks forward towards some consequences of our model for learning and interpretation. We then draw some conclusions in Section \ref{sec:conclusions}.   A detailed and formal analysis of interpretive bias has important social implications.  Questions of bias are not only timely but also  pressing for democracies that are having a difficult time dealing with campaigns of disinformation and a society whose information sources are increasingly fragmented and whose biases are often concealed.  Understanding linguistic and cognitive mechanisms for bias precisely and algorithmically can yield valuable tools for navigating in an informationally bewildering world.   

\section{The model of interpretive bias} \label{sec:model}

As mentioned in Section \ref{sec:aims}, understanding interpretive bias requires two ingredients.  First, we need to know what it is to interpret a text or to build a history over a set of facts.    Our answer comes from analyzing discourse structure and interpretation in SDRT \cite{asher:1993,asher:lascarides:2003}.   A history for a text connects its elementary information units, units that convey propositions or describe events, using semantic relations that we call {\em discourse relations} to construct a coherent and connected whole.   Among such relations are logical, causal, evidential, sequential and resemblance relations as well as relations that link one unit with an elaboration of its content.  It has been shown in the literature that discourse structure is an important factor in accurately extracting sentiments and opinions from text \cite{cadilhac:etal:2011,cadilhac:etal:2012a,cadilhac:etal:2013}, and our examples show that this is the case for interpretive bias as well.  
\subsection{Epistemic ME games}
The second ingredient needed to understand interpretive bias is the connection between on the one hand the purpose and assumption behind telling a story and on the other the particular way in which that story is told.
A history puts the entities to be understood into a structure that serves certain purposes or conversational goals \cite{grosz:sidner:1986}.  Sometimes the history attempts to get at the ``truth'', the true causal and taxonomic structure of a set of events.   But a history may also serve other purposes---e.g., to persuade, or to dupe an audience.  
Over the past five years,    \cite{JPL,IACL, LACL,SEMDIAL} have developed an account of conversational purposes or goals and how they guide strategic reasoning in a framework called {\em Message Exchange (ME) Games}.  ME games provide a general and formally precise framework for not only the analysis of conversational purposes and conversational strategies, but also for the typology of dialogue games from \cite{walton:1984} and finally for the analysis of strategies for achieving what we would intuitively call ``unbiased interpretation'', as we shall see in the next section.  In fact in ME Games, conversational goals are analyzed as properties, and hence sets, of conversations; these are the conversations that ``go well'' for the player.   ME games bring together the linguistic analysis of SDRT with a game theoretic approach to strategic reasoning; in an ME game, players alternate making sequences of discourse moves such as those described in SDRT, and a player wins if the conversation constructed belongs to her winning condition, which is a subset of the set of all possible conversational plays.  ME games are designed to analyze the interaction between conversational structure, purposes and assumptions, in the absence of  assumptions about cooperativity or other cognitive hypotheses, which can cause problems of interpretability in other frameworks \cite{venant:2016}.   ME games also assume a Jury that sets the winning conditions and thus evaluates whether the conversational moves made by players or conversationalists are successful or not.  The Jury can be one or both of the players themselves or some exogenous body.  

To define an ME game, we first fix a finite set of {\sf players} $N$ and let $i$ range over $N$. For simplicity, we consider here the case where there are only two players, that is $N=\{0,1\}$, but the notions can be easily lifted to the case where there are more than two players. Here, Player $(1-i)$ will denote the opponent of Player $i$. We need a vocabulary $V$ of moves or actions; these are the discourse moves as defined by the language of SDRT.  The intuitive idea behind an ME game is that a conversation proceeds in {\sf turns} where in each turn one of the players `speaks' or plays a string of elements from $V$. In addition, in the case of conversations, it is essential to keep track of ``who says what''. To model this, each player $i$ was assigned a copy $V_i$ of the vocabulary $V$ which is simply given as $V_i=V\times\{i\}$.   As \cite{JPL} argues, a conversation may proceed indefinitely, and so conversations correspond to {\sf plays} of ME games, typically denoted as $\play$, which are the union of finite or infinite sequences in $\vocab$, denoted as $\vocab^*$ and $\vocab^\omega$ respectively. The set of all possible conversations is thus $(\vocab^*\cup \vocab^\omega)$ and is denoted as $\vocab^\infty$.
\begin{definition}[ME game \cite{JPL}]
  A {\sf Message Exchange game (ME game)}, $\game$, is a tuple $((V_0 \cup V_1)^\infty, \jury)$ where $\jury$ is a Jury.
\end{definition}
Due to the ambiguities in language, discourse moves in SDRT are underspecified formulas that  may yield more than one fully specified discourse structure or {\em histories} for the conversation; a resulting play in an ME game thus forms one or more {\em histories} or complete discourse structures for the entire conversation.

To make ME games into a truly realistic model of conversation requires taking account of the limited information available to conversational participants.  \cite{JOLLI} imported the notion of a type space from epistemic game theory \cite{harsanyi:1967} to take account of this.  The {\sf type} of a player $i$ or the Jury is an abstract object that is used to code-up anything and everything about $i$ or the Jury, including her behavior, the way she strategizes, her personal biases, etc. \cite{harsanyi:1967}. 
  Let $\Strat_i$ denote the set of strategies for Player $i$ in an ME game; let $\Strat=\Strat_0\times\Strat_1$; and let $\Strat^\play_{i}$ be the set of strategies of $i$ given play $\play$.
  \begin{definition}[Harsanyi type space \cite{harsanyi:1967}] A Harsanyi {\sf type space} for $\Strat$ is a tuple $\Type=(\{\type_i\}_{i\in\{0,1\}},\type_\jury,\{\bbelief^\play_i\}_{i\in\{0,1\},\play\in\Play},\{\bbelief^\play_\jury\}_{\play\in\Play},\Strat)$ such that $\type_\jury$ and $\type_i$, for each $i$, are non-empty (at-most countable) sets called the {\sf Jury-types} and {\sf $i$-types} respectively and $\{\bbelief^\play_i\}$ and $\{\bbelief^\play_\jury\}$ are the beliefs of Player $i$ and the Jury respectively at play $\play\in\Play$.  
\end{definition}
\cite{JOLLI} defines the beliefs of the players and Jury using the following functions.
\begin{definition}[Belief function] For every play $\play\in\Play$ the {\sf (first order) belief} $\bbelief_i^\play$ of player $i$ at $\play$ is a pair of measurable functions $\bbelief^\play_i = (\belief^\play_i,\intpr^\play_i)$ where $\belief^\play_i$ is the {\sf belief function} and $\intpr^\play_i$ is the {\sf interpretation function} defined as:
  $$\belief^\play_i: \type_i \times \hist(\play) \rightarrow \Delta(\type_{(1-i)}\times \Strat^\play_{(1-i)}\times \type_\jury)$$
$$\intpr^\play_i: \type_i \times \type_{(1-i)} \times \type_\jury \rightarrow \Delta(\hist(\play))$$

where $\Delta(\cdot)$ is the set of probability distributions over the corresponding set. Similarly the (first order) belief $\bbelief^\play_\jury$ of the Jury is a pair of measurable functions $\bbelief^\play_\jury=(\belief^\play_\jury,\intpr^\play_\jury)$ where the {\sf belief function} $\belief^\play_i$ and the {\sf interpretation function} $\intpr^\play_i$ are defined as:
$$\belief^\play_\jury: \type_\jury \times \hist(\play) \rightarrow \Delta(\type_0\times\Strat^\play_0\times\type_1\times \Strat^\play_1)$$
$$\intpr^\play_\jury: \type_\jury \times \type_0 \times \type_1 \rightarrow \Delta(\hist(\play))$$
\end{definition}
Composing $\belief$ and $\intpr$ together over their respective outputs reveals a correspondence between interpretations of plays and types for a fixed Jury type $\tau$: every history yields a distribution over types for the players and every tuple of types for the players and the Jury fixes a distribution over histories.  We'll call this the {\em types/history correspondence}.

An epistemic ME game is an ME game with a Harsanyi type space and a type/history correspondence as we've defined it. By adding types to an ME game, we provide the beginnings of a game theoretic model of interpretive bias that we believe is completely new.   Our definition of bias is now:
\begin{definition} [Interpretive Bias] An interpretive bias in an epistemic ME game is the probability distribution over types given by the belief function of the conversationalists or players, or the Jury.
\end{definition}
Note that in an ME game there are typically several interpretive biases at work: each player has her own bias, as does the Jury. 

 Outside of language, statisticians study bias;  and sample bias is currently an important topic.\footnote{See https://www.elen.ucl.ac.be/esann/index.php?pg=specsess\#biasesbigdata.}  To do so, they exploit statistical models with a set of parameters and random variables, which play the role of our types in interpretive bias.  
But for us, the interpretive process is already well underway once the model, with its constraints, features and explanatory hypotheses, is posited; at least a partial history, or set of histories, has already been created.

The ME model in \cite{JOLLI} not only makes histories dependent on biases but also conditionally updates an agent's bias, the probability distribution, given the interpretation of the conversation or more generally a course of events as it has so far unfolded and crucially as the agent has so far interpreted it.  This means that certain biases are reinforced as a history develops, and in turn strengthen the probability of histories generated by such biases in virtue of the types/histories correspondence.  We now turn to an analysis of \ref{Sheehan} discussed in \cite{JPL,JOLLI} where arguably this happens.

\subsection{Analyzing the Sheehan example}\label{subsec:complex}


\hidden{As background to this excerpt from a press conference by Senator Coleman's spokesman Sheehan, Senator Coleman was running for re-election as a senator from Minnesota in the 2008 US elections.

\begin{enumerate}[label=S.\alph*]
\item {\bf Reporter:} {\myfont On a different subject is there a reason that the Senator won't say whether or not someone else bought some suits for him?}\label{Sa}
\item {\bf Sheehan:} {\myfont Rachel, the Senator has reported every gift he has ever received.}\label{Sb} 
\item {\bf Reporter:} {\myfont That wasn't my question, Cullen.}\label{Sc} 
\item {\bf Sheehan:} (i) {\myfont The Senator has reported every gift he has ever received.} (ii) {\myfont We are not going to respond to unnamed sources on a blog.}\label{Sd} 
\item {\bf Reporter:} {\myfont So Senator Coleman's friend has not bought these suits for him? Is that correct?}\label{Se} 
\item {\bf Sheehan:} {\myfont The Senator has reported every gift he has ever received.}\label{Sf} 
\end{enumerate}
Sheehan continues to repeat, {\myfont ``The Senator has reported every gift he has ever received''} seven more times in two minutes to every follow up question by the reporter corps. {\tt http://www.youtube.com/watch?v=VySnpLoaUrI}
\medskip
}
To formulate \ref{Sheehan} as an ME game, \cite{JOLLI} assumes  two active players (i) the reporter corps (R) and (ii) spokesman Sheehan (S).   The play in is given in \ref{Sheehan}.

We assume our players commit to the unambiguous contents of their discourse moves.  To elucidate this assumption,  we note first that there are many relevant parameters to determining discourse structure---for instance, the various features used by discourse parsing models \cite{perret:etal:2016}, whose probability of supporting a particular discourse structure for a text is learned from a corpus.  But for now we assume that this information, when it is encoded in the lexicon or the grammar, is common knowledge and part of a speaker's competence.  

But in \ref{Sheehan} there are also moves with an ambiguous or underspecified content, and those are the turns on which Sheehan uses the phrase, {\em the Senator has reported every gift he has ever received}.   There are choices as to what the discourse function of that phrase is---e.g., is Sheehan answering the reporter's question for instance or evading it?     So we need to make a distinction between an {\em underspecified logical form} for the dialogue or \ulf~ \cite{asher:lascarides:2003} and a full history, as explained in \cite{JOLLI}.   
Let's call the underspecified play of explicit commitments for \ref{Sheehan} $\play$.


\hidden{
$\play$ serves as the spine of the game tree depicted in Figure \ref{fig:SheehanME}.
The relation instances whose arguments are underspecified are drawn in red.
\begin{figure}
\input{SheehanMEgame}
\caption{An ME game structure for the initial part of the conversation between R and S. The nodes represent the EDUs and the rectangular boxes represent the CDUs as given in the description of $\rho$.}\label{fig:SheehanME}
\end{figure}
}
To analyze \ref{Sheehan} \cite{JOLLI} assumes two types for Sheehan that are relevant to interpreting what he said: $\tp_H$ and $\tp_D$.\footnote{These two types to illustrate how types can affect interpretation.  In reality, there might be many more.}  
\begin{itemize}
\item $\tp_H$ is the `honest' type, according to which Sheehan truly implicates that the Senator did not receive the suits and that he simply does not want to respond to this charge.
\item $t_D$ is the `dishonest' type, according to which Coleman received the suits but did not declare them and Sheehan is trying to cover this fact up.
\end{itemize}

\cite{JOLLI} also envisages two types for the Jury: $\tj_U$ and $\tj_B$.
\begin{itemize}
\item $\tj_U$ is the `unbiased' Jury that starts out with a presumption of full disclosure and honesty from Sheehan but reserves judgment about whether the Senator received the suits.
\item $\tj_B$ is the type of the Jury that is disposed to believe in Sheehan's confirmation of the Senator's innocence.  Such a Jury is `biased', in virtue of its prior beliefs that dispose it to interpret Sheehan's moves, regardless of what they are, in a favorable light.   
\end{itemize}

These two types influence how the underspecified elements in $\play$ are interpreted.  In particular, 
two different interpretations of the underspecified elements give rise to two different histories, as detailed in \cite{JOLLI}.   
The types $\tj_U$ and $\tj_B$ of the Jury have different priors concerning S's types and this is what drives their differing interpretations.  $\tj_U$ starts with an indifference between $\tp_H$ and $\tp_D$, while $\tj_B$ starts off believing S is of type $\tp_H$, with a high probability.
\hidden{
\begin{table}
  \renewcommand\thetable{4a}
\begin{center}
\begin{tabular}{|c|c|c|}
\hline
$\belief^{\play_0}_\jury(\ \cdot\ ,\epsilon)$ & $\tp_H$ & $\tp_D$  \\
\hline
$\tj_U$ & 0.5 & 0.5 \\
\hline
$\tj_B$ & 0.7 & 0.3 \\
\hline
\end{tabular}
\qquad\qquad
\begin{tabular}{|c|c|c|}
\hline
$\belief^{\play_j}_\jury(\cdot)$ & $\tp_H$ & $\tp_D$  \\
\hline
$\hist^1$ & 0.5 & 0.5 \\
\hline
$\hist^2$ & 0.7 & 0.3 \\
\hline
\end{tabular}
\end{center}
\caption{The beliefs of the Jury about the type of S before the start of the game and the beliefs of the Jury about the type of S throughout the play $\play$}
\label{tab:beliefJ0}
\end{table}
Now As the play $\play$ progresses, the types $\tj_U$ and $\tj_B$ of the Jury end up interpreting $\play$ differently.}  \cite{JOLLI} gives a detailed analysis of how the two different interpretations come about from different beliefs.  But in addition, it shows how these interpretations change the beliefs of the two Juries about the honesty of Sheehan; guided by its interpretation of $\play$ and using Bayesian conditionalization, $\tj_U$ comes eventually to believe that Sheehan is of type $\tp_D$, while   $\tj_B$ uses the same techniques but crucially a different interpretation to reinforce its belief that Sheehan is of type $\tp_H$. 

\subsection{Generalizing from the case study}
The Sheehan case study in \cite{JOLLI} shows the interactions of interpretation and probability distributions over types.  We'll refer to content that exploit assumptions about types' {\em epistemic content}.  \ref{Sheehan} also offers a case of a self-confirming bias with Jury $\tj_B$.\footnote{Self-reinforcing biases of nonlinguistic facts are also echoed in popular analyses, for instance `The Evangelical Roots of Our Post-Truth Society' by Molly Worthen, {\em New York Times}, 16.04.2017. But as far as we know, only \cite{JOLLI} has provided at least a partial formal analysis of this phenomenon.}  But the analysis proposed by \cite{JOLLI} leaves open an important open question about what types are relevant to constructing a particular history and only examines one out of many other cases of biased interpretation.  In epistemic game models, the relevant types are typically given exogenously and Harsanyi's type space construction is silent on this question.  The question seems {\em a priori} very hard to answer, because anything and everything might be relevant to constructing a history.  

 In \ref{Sheehan}, the relevant types have to do with the interpreters' or Jurys' attitudes towards the commitments of the spokesman and Coleman.  These attitudes might reinforce or be a product of other beliefs like beliefs about the spokesman's political affiliations.  But we will put forward the following simplifying hypothesis:\\
 
 \medskip
 \noindent
{\bf Hypothesis 1}: epistemic content is based on assumptions about types defined by different attitudes to commitments by the players and or the Jury to the contents of a discourse move or sequence of discourse moves.\\

\medskip
\noindent
{\bf Hypothesis 2}:  These assumptions
can be represented as probability distributions over types. \\

\medskip
\noindent
In \ref{Sheehan}, we've only looked at epistemic content from the point of view of the interpreter, which involves types for the Jury defined in terms of probability distributions over types for the speaker.   But we can look at subjective interpretations from the perspective of the speaker as well.  In other words, we look at how the speaker might conceptualize the discourse situation, in particular her audience.  We illustrate this with another type of content based on types.    Consider the following move by Marion Le Pen, a leader of the French nationalist, right-wing party {\em le Front National} in which she recently said: 
\ex. \label{lepen} La France \'etait la fille a\^in\'ee de l'\'eglise. Elle est en passe de devenir la petite ni\`ece de l'Islam. (France was once the eldest daughter of the Catholic church. It is now becoming the little niece of Islam.)\footnote{https://www.youtube.com/watch?v=9r8fKymWWZ8}

\ref{lepen} appeals to what the speaker takes to be her intended audience's beliefs about Islam, Catholicism and France.  In virtue of these beliefs, 
this discourse move takes on a loaded racist meaning, conveying an assault on France and its once proud status by people of North African descent.   Without those background beliefs, however, Le Pen's statement might merely be considered a somewhat curious description of a recent shift in religious majorities.   This is known as a ``dog whistle,'' in which a discourse move communicates a content other than its grammatically determined content to a particular audience \cite{henderson:mccready:2017}.  While \cite{stanley:2015} proposes that such messages are conventional implicatures, \cite{henderson:mccready:2017,khoo:2017} show that dog whistle content doesn't behave like other conventional implicatures; in terms of tests about ``at issue content'', dog whistle content patterns with other at issue content, not with the content associated with conventional implicatures in the sense of \cite{potts:2005}.  This also holds of content that resolves ambiguities as in \ref{Sheehan}.   

The dogwhistle content seems to be driven by the hearer's type in \ref{lepen} or the speaker's beliefs about the interpreter's or hearer's type.   Generalizing from \cite{burnett:2017}, the use of the historical expression {\em la fille ain\'ee de l'\'eglise} contrasted with {\em la petite ni\`ece} has come to encode a type, in much the same way that dropping the final {\em g} in present participles and gerunds has come to signify a type \cite{burnett:2017}, for the speaker $i$ about hearer $j$; e.g.,  $j$ will believe that $i$ has the strategy of using just this language to access the loaded interpretation and moreover will identify with its content.  Because this meaning comes about in virtue of the hearer's type, the speaker is in a position to plausibly deny that they committed to conveying a racist meaning, which is a feature of such dog whistles.\footnote{\cite{henderson:mccready:2017} appeal to an epistemic signaling game approach like that of \cite{burnett:2017} to talk about communicating that a speaker is of a certain type and believes that the interpreter is of that type too.   However, the types in \cite{burnett:2017} have to do with politeness registers, not generalized beliefs about interpretation.} 
In fact, we might say that all dogwhistle content is so determined.

We can complicate the analysis by considering the speaker's types, the interlocutor's types and types for the Jury when these three components of an ME game are distinct (i.e. the Jury is distinct from the interlocutors).   A case like this is the Bronston example discussed in \cite{JOLLI}.

By looking at dogwhistles, we've now distinguished two kinds of epistemic content that depends on an interpreters' type.  The epistemic content may as in \ref{Sheehan} fill out the meaning of  an underspecified play to produce a determinate history.\footnote{Note that this is a precise epistemic game-theoretic model of what \cite{recanati:2004}, for example, called "pragmatic saturation."  However, in following the ideas of \cite{burnett:2017}, we should rather call this ``sociolinguistic saturation.''}  Dog whistles add content to a specific discourse unit that goes beyond its grammatically determined meaning.\footnote{This in turn is a precise model of what called relevance theorists like \cite{sperber:wilson:1986} have called "free enrichment."}  More formally, we can define these two kinds of epistemic content using the machinery of ME games.  Given that plays in an ME game are sequences of discourse moves, we can appeal to the semantics of these moves and a background consequence relation $\models$ defined as usual.  In addition, a play $\play$ in an ME game may itself be a fully specified history or a sequence of discourse moves that is compatible with several fully specified histories given a particular interpreter's or Jury's type $J$.  Let $\hist(\play)_J$ be the set of histories (FLFs) compatible with a play $\play$ given an interpreter or Jury type $J$. $\play$ will be ambiguous and open to epistemic content supplementation just in case: (i) $|\hist(\play)|_J > 1$ for any type $J$ for a linguistically competent jury, and (ii) there are $\hist_1, \hist_2 \in \hist(\play)_J$, such that $\hist_1$ and $\hist_2$ are semantically distinct (neither one entails the other).  Now suppose that a play $\play$ gives rise through the grammar to a history, $\hist^*_\play$. Then $\play$ is a {\em dog whistle} for $J$ just in case: (i) $|\hist(\play)|_J > 1$, (ii) $\hist^*_\play\in \hist(\play)$ and (iii) there is a $\hist^\dagger_\play\in \hist(\play)_J$ that can positively affect some jury perhaps distinct from $J$ and such that $\hist^\dagger_\play \models \hist^*_\play$.  On this definition, a player who utters such a play $\play$ always has the excuse that what he/she actually meant was $\hist^*_\play$ when challenged---which seems to be one essential feature of a dog whistle.
  
Plays with such semantic features may not be a pervasive feature of conversation; not every element is underspecified or is given a content over and above its linguistically determined one.  But in interpreting a set of nonlinguistic facts ${\cal X}$ or data not already connected together in a history, that is in constructing a history over ${\cal X}$, an interpreter $i$, who in this case is a speaker or writer, must appeal to her beliefs, which includes her beliefs about the Jury to whom her discourse actions are directed.  So certainly the type of $i$, which includes beliefs about the Jury for the text, is relevant to what history emerges.  The facts in ${\cal X}$ don't wear their relational properties to other facts on their sleeves so to speak, and so $i$ has to supply the connections to construct the history.  In effect for a set of non linguistically given facts, ``ambiguities of attachment,'' whose specification determines how the facts in ${\cal X}$ are related to each other, are ubiquitous and must be resolved in constructing a history.  The speaker or ``history creator'' $i$'s background beliefs  determine the play and the history an interpreter $j$ takes away.     

In the case of constructing a history over a set of nonlinguistic facts ${\cal X}$, the interpreter $j$'s task of getting the history $i$ has constructed will not reliably succeed unless one of two conditions are met: either $i$ and $j$ just happen to share the relevant beliefs (have close enough types) so that they construct the same histories from ${\cal X}$, or $i$ uses linguistic devices to signal the history.\footnote{the space of possibilities is too vast for $j$ to converge on a determinate history representing the commitments of the speaker $i$ with any confidence, if $X$ has any complexity and $i$ doesn't give any hint about the history she commits to over $X$.  ME games require that the relevant types for the interpretation of a conversation affect conversational strategies and the winning conditions of the players.   By itself, however, our assumption doesn't delimit the set of relevant types to a manageable set.  The set of all possible winning conditions and strategies is much larger than what a Harsanyi type space for an ME game allows; recall, it allows an at most countable set of types for each player and the Jury, but the space of all conversations has cardinality $\aleph_1$, so using the Generalized Continuum Hypothesis the set of all possible winning conditions has cardinality $\aleph_2$.}   ME games require winning conversations, and by extension texts, to be (mostly) {\em coherent}, which means that the discourse connections between the elements in the history must be largely determined in any successful play, or can be effectively determined by $j$.  This means that $i$ will usually reveal relevant information about her type through her play, in virtue of the type/history correspondence, enough to reconstruct the history or much of it.  In the stories on the March for Science, for example, the reporters evoke very different connections between the march and other facts.  The {\em Townhall} reporter, for instance, connects the March for Science to the Women's march and ``leftwing'' political manifestations and manifests a negative attitude toward the March.  But he does so so unambiguously that little subjective interpretation on the part of the interpreter or Jury is needed to construct the history or assign a high probability to a type for $i$ that drives the story.   


This discussion leads to the following observations.  To construct a history over a set of disconnected nonlinguistic facts $\mathcal{X}$, in general a Jury needs to exploit linguistic pointers to the connections between elements of $\mathcal{X}$, if the speaker is to achieve the goal of imparting a (discourse) coherent story, unless the speaker knows that the Jury or interpreter has detailed knowledge of her type.  The speaker may choose to leave certain elements underspecified or ambiguous, or use a specified construction, to invoke epistemic content for a particular type that she is confident the Jury instantiates.  
How much so depends on her confidence in the type of the Jury.  This distribution or confidence level opens a panoply of options about the uses of epistemic content: at one end there are histories constructed from linguistic cues with standard, grammatically encoded meanings; at the other end there are histories generated by a code shared with only a few people whose types are mutually known.   As the conversation proceeds as we have seen, probabilities about types are updated and so the model should predict that a speaker may resort to more code-like messages in the face of feedback confirming her hypotheses about the Jury's type (if such feedback can be given) and that the speaker may revert to a more message exploiting grammatical cues in the face of feedback disconfirming her hypotheses about the Jury's type.  Thus, the epistemic ME model predicts a possible change in register as the speaker receives more information about the Jury's type, though this change is subject to other conversational goals coded in the speaker's victory condition for the ME game.   

\subsection{ME persuasion games} 
We've now seen how histories in ME games bring an interpretive bias, the bias of the history's creator, to the understanding of a certain set of facts.  We've also seen how epistemic ME games allow for the introduction of epistemic content in the interpretation of plays.  Each such epistemic interpretation is an instance of a bias that goes beyond the grammatically determined meaning of the play and is dependent upon the Jury's  or interpreter's type.   We now make explicit another crucial component of ME games and their relation to bias: the players' winning conditions or discourse goals.  Why is this relevant to a study of bias?  The short answer is that players' goals tells us whether two players' biases on a certain subject are compatible or resolvable or not.  Imagine that our two Juries in \ref{Sheehan} shared the same goal---of getting at the truth behind the Senator's refusal to comment about the suits.  They might still have come up with the opposing interpretations that they did in our discussion above.  But they could have discussed their differences, and eventually would have come to agreement, as we show below in Proposition \ref{agree}.\footnote{Note that our result extends the famous result of \cite{aumann:1976}, as we are not assuming common priors, but only a rational process of belief revision in the face of evidence.}

However, our two Juries might have different purposes too.  One Jury might have the purpose of finding out about the suits, like the reporters; the other might have the purpose just to see Senator Coleman defended, a potentially quite different winning condition and collection of histories.  In so doing we would identify Jury 1 with the reporters or at least Rachel, and Jury 2 with Sheehan.  Such different discourse purposes have to be taken into account in attempting to make a distinction between good and bad biases.  From the perspective of subjective rationality or rationalizability (an important criterion in epistemic game theory \cite{battigalli:2003}), good biases for a particular conversation should be those that lead to histories in the winning condition, histories that fulfill the discourse purpose; bad biases lead to histories that do not achieve the winning condition.  The goals that a Jury or interpreter $j$ adopts and her biases go together; $j$'s interpretive bias is good for speaker $i$, if it helps $i$ achieve her winning condition.  Hence, $i$'s beliefs about $j$ are crucial to her success and rationalizable behavior.  Based on those beliefs $i$'s behavior is rationalizable in the sense we have just discussed.  If she believes Jury 2 is the one whose winning condition she should satisfy, there is no reason for her to change that behavior.   Furthermore, suppose Jury 1 and Jury 2 discuss their evaluations; given that they have different goals, there is no reason for them to come to an agreement with the other's point of view either.  Both interpretations are rationalizable as well, if the respective Juries have the goals they do above.  A similar story applies to constructing histories over a set of facts, in so far as they had different conceptions of winning conditions set by their respective Juries.  In contrast to Aumann's dictum \cite{aumann:1976}, in our scenario there is every reason to agree to disagree!\footnote{Technically, Aumann's observation relies on common prior probabilities.  We don't see any reason to adopt such an assumption in an analysis of strategic conversations or bias.  Our observation is sort of a correlate or converse of Aumann's.}




Understanding such discourse goals is crucial to understanding bias for at least two reasons.  The first is that together with the types that are conventionally coded in discourse moves, they fix the space of relevant types.  In \ref{Sheehan}, 
Jury 1 is sensitive to a winning condition in which the truth about the suits is revealed, what we call a {\em truth oriented goal}. The goal of Jury 2, on the other hand, is to see that Coleman is successfully defended, what we call a {\em persuasion goal}.  In fact, we show below that a truth oriented goal is a kind of persuasion goal.  Crucial to the accomplishment of either of these goals is for the Jury $j$  to decide whether the speaker $i$ is committing to a definite answer that she will defend (or better yet an answer that she believes) on a given move to a question from her interlocutor or is $i$ trying to avoid any such commitments.  If it's the latter, then $j$ would be epistemically rash to be persuaded.  But the two possibilities are just the two types for Sheehan that are relevant to the interpretation of the ambiguous moves in \ref{Sheehan}. Because persuasive goals are almost ubiquitous at least as parts of speaker goals, not only in conversation but also for texts (think of how the reporters in the examples on the March for Science are seeking to convince us of a particular view of the event), we claim that these two types 
are relevant to the interpretation of many, if not all, conversations.   In general we conjecture that the relevant types for interpretation may all rely on epistemic requirements for meeting various kinds of conversational goals.

The second reason that discourse goals are key to understanding bias is that by analyzing persuasion goals in more detail we get to the heart of what bias is.  Imagine a kind of ME game played between two players, E(lo\"ise) and A(belard), where E proposes and tries to defend a particular interpretation of some set of facts ${\cal X}$, and A tries to show the interpretation is incorrect, misguided, based on prejudice or whatever will convince the Jury to be dissuaded from adopting E's interpretation of ${\cal X}$.  As in all ME games, E's victory condition in an ME persuasion game is a set of histories determined by the Jury, but but it crucially depends on E's and A's beliefs about the Jury:  E has to provide a history $\hist$ over ${\cal X}$; A has to attack that history in ways that accord with her beliefs about the Jury; and E has to defend $\hist$ in ways that will, given her beliefs, dispose the Jury favorably to it.

  
 An ME persuasion game is one where E and A each present elements of ${\cal X}$ and may also make argumentative or attack moves in their conversation.  At each turn of the game, A can argue about the history constructed by E over the facts given so far, challenge it with new facts or attack its assumptions, with the result that E may rethink and redo portions her history over ${\cal X}$ (though not abandon the original history entirely) in order to render A's attack moot.  E wins if the history she finally settles on for the facts in $X$ allows her to rebut every attack by A; A wins otherwise.  
   
 A reasonable precisification of this victory condition is that the proportions of good unanswered attacks on the latest version of E's history with respect to the total number of attacks at some point continues to diminish and eventually goes to $0$.  This is a sort of limit condition: if we think of the initial segments $n$ E's play as producing an ``initial'' history $\hist_n^E$ over ${\cal X}$, as  $n \rightarrow \infty$, $\hist_n^E$ has no unanswered counterattacks by A that affect the Jury.
\hidden{
 The complexity of this winning condition can be characterized using the mathematical structure of an ME game \cite{JPL}.  More precisely, where a {\em good attack} by A is one to which E does not have a convincing response for the Jury and $p_n$ is a prefix of play $p$ of length $n$, E's winning condition is:
\ex. \label{win1}
  $ \win_E \ = \ \bigcup\{p \in (V_E \cup V_A)^\infty \ | \ \limsup_{n \rightarrow \infty} \frac{\mbox {good attacks by } A \mbox{ in } p_n}{\mbox{attacks by } A \mbox{ in } p_n} = 0\}$

 $\win_E$ can be transformed to:\\
   $$\win_E \ = \ \{p \in (V_E \cup V_A)^\infty \ | \ \limsup_{n \rightarrow \infty} \frac{\mbox {good attacks by } A \mbox{ in } p_n}{\mbox{attacks by } A \mbox{ in } p_n} = 0\}$$

 \begin{align*}
   \win_E  &= \bigcap_{N>0}\left\{\play\in(V_E \cup V_A)^\infty\ \mid\ \frac{\mbox {good attacks by } A \mbox{ in } p_n}{\mbox{attacks by } A \mbox{ in } p_n}  < 1/N\right\}\\
   &= \bigcap_{N>0}\bigcup_{m>0}\bigcap_{n\geq m}\left\{\play\in(V_E \cup V_A)^\infty\ \mid\ \frac{\mbox {good attacks by } A \mbox{ in } p_n}{\mbox{attacks by } A \mbox{ in } p_n}  < 1/N\right\}
 \end{align*}

Now note that the set
$$\bigcap_{n\geq m}\left\{\play\in(V_E \cup V_A)^\infty\ \mid\ \frac{\mbox {good attacks by } A \mbox{ in } p_n}{\mbox{attacks by } A \mbox{ in } p_n}  < 1/N\right\}$$
\noindent
is closed. So, $\win_E$ is a $\Pi^0_3$ set.}  Such winning histories are extremely difficult to construct; as one can see from inspection, no finite segment of an infinite play guarantees such a winning condition.\footnote{The complexity of this winning condition can be characterized using the mathematical structure of an ME game, in which winning conditions can be characterized in terms of the Borel hierarchy.   For details see \cite{JPL}.}
We shall call a history segment that is part of a history in $E$'s winning condition as we have just characterized it, {\em E-defensible}.

The notion of an ME persuasion game opens the door to a study of attacks, a study that can draw on work in argumentation and game theory \cite{dung:1995,glazer:rubinstein:2004,besnard:hunter:2008}.  
ME games and ME persuasion games in particular go beyond the work just cited, however, because our notion of an effective attack involves the type of the Jury as a crucial parameter; the effectiveness of an attack for a Jury relies on its prejudices, technically its priors about the game's players' types (and hence their beliefs and motives).    For instance, an uncovering of an agent's racist bias when confronted with a dog whistle like that in \ref{lepen} is an effective attack technique if the respondent's type for the Jury is such that it is sensitive to such accusations, while it will fail if the Jury is insensitive to such accusations.  ME games make plain the importance in a persuasion game of accurately gauging the beliefs of the Jury!

\subsection{ME truth games}

We now turn to a special kind of ME persuasion game with what we call a {\em disinterested} Jury.  The intuition behind a disinterested Jury is simple: such a Jury judges the persuasion game based only on the public commitments that follow from the discourse moves that the players make.  It is not predisposed to either player in the game.  While it is difficult to define such a disinterested Jury in terms of its credences, its probability distribution over types, we can establish some necessary conditions.  We first define the notion of the {\sf dual} of a play of an ME game. Let $(v,i)\in \mevocab$ be an element of the labeled vocabulary with player $i \in \{0,1\}$. Define its dual as:
$$\overline{(v,i)} = (v,1-i)$$
The dual of a play $\play\in(\mevocab)^\infty$ then is simply the lifting of this operator over the entire sequence of $\play$. That is, if $\play=x_0x_1x_2\ldots$, where $x_0=\epsilon$ then
$$\overline{\play} = x_0\overline{x_1}\ \overline{x_2}\ldots$$

Then, a disinterested Jury must necessarily satisfy:

\begin{itemize}
\item {\bf Indifference towards player identity:} A Jury $\jury=(\win_0,\win_1)$ is unbiased only if for every $\play\in \mevocab$, $\play\in \win_i$ iff $\overline{\play}\in \win_{(1-i)}$.
\item {\bf Symmetry of prior belief:} A Jury is unbiased only if it has symmetrical prior beliefs about the player types. 
\end{itemize}
Clearly, the Jury $\tj_B$ does not have symmetrical prior beliefs nor is it indifferent to player identity, while Jury $\tj_U$ arguably has symmetrical beliefs about the participants in \ref{Sheehan}.  Note also that while Symmetry of prior beliefs is satisfied by a uniform distribution over all types, but it does not entail such a uniform distribution.  Symmetry is closely related to the principle of maximum entropy used in fields as diverse as physics and computational linguistics\cite{berger:etal:1996}, according to which in the absence of any information about the players would entail a uniform probability distribution over types.

A distinterested Jury should evaluate a conversation based solely on the strength of the points put forth by the participants.  But also crucially it should  evaluate the conversation in light of the {\em right } points.  So for instance, appeals to ad hominem attacks by A or colorful insults should not sway the Jury in favor of A.  They should evaluate only based on how the points brought forward affect their credences under conditionalization.  A distinterested Jury is impressed only by certain attacks from A, ones based on evidence (E's claims aren't supported by the facts) and on formal properties of coherence, consistency and explanatory or predictive power.    In such a game it is common knowledge that attacks based on information about E's type that is not relevant either to the evidential support or formal properties of her history are ignored by the Jury and the participants know this.  The same goes for E; counterattacks by her on A that are not based on evidence or the formal properties mentioned above. 

\cite{JPL} discusses the formal properties of coherence and consistency in detail, and we say more about explanatory and predictive power below.  The evidential criterion, however, is also particularly important, and it is one that a disinterested Jury must attend to.  Luckily for us, formal epistemologists have formulated constraints like {\em cognitive skill} and {\em safety} or {\em anti-luck} on beliefs that are relevant to characterizing this evidential criterion \cite{moss:2013,konek:2016}.  Cognitive skill is a factor that affects the success (accuracy) of an agent's beliefs:  the success of an agent's beliefs is the result of her cognitive skill, exactly to the extent that the reasoning process that produces them makes evidential factors (how weighty, specific, misleading, etc., the agent's evidence is) comparatively important for explaining that success, and makes non-evidential factors comparatively unimportant.  In addition, we will require that the relevant evidential factors are those that have been demonstrated to be effective in the relevant areas of inquiry.  So if a Jury measures the success of a persuasion game in virtue of a criterion of cognitive ability on the part of the participants and this is common knowledge among the participants (something we will assume throughout here), then, for instance, A's attacks have to be about the particular evidence adduced to support E's history, the way it was collected or verifiable errors in measurements etc., and preclude general skeptical claims from credible attacks in such a game.  These epistemic components thus engender more relevant types for interpretation: are the players using cognitive skill and anti-luck conditions or not?  More particularly, most climate skeptics' attacks on climate change science, using general doubts about the evidence without using any credible scientific criteria attacking specific evidential bases, would consequently be ruled as irrelevant in virtue of a property like cognitive skill.  But this criterion may also affect the Jury's interpretation of the conversation. A Jury whose beliefs are constrained by cognitive ability will adjust its beliefs about player types and about interpretation only in the light of relevant evidential factors. 

{\em Safety} is a feature of beliefs that says that conditionalizing on circumstances that could have been otherwise without one's evidence changing should not affect the strength of one's beliefs.  Safety rules out out belief profiles in which luck or mere hunches play a role.

The notion of a disinterested jury is formally a complicated one.  Consider an interpretation of a conversation between two players E and A.   Bias can be understood as a sort of modal operator over an agent's first order and higher order beliefs.  So a disinterested Jury in an ME game means that neither its beliefs about A nor about E involve an interested bias; nor do its beliefs about A's beliefs about E's beliefs or E's beliefs about the A's beliefs about E's beliefs, and so on up the epistemic hierarchy.  Thus, a disinterested Jury in this setting involves an infinitary conjunction of modal statements, which is intuitively (and mathematically) a complex condition on beliefs.  And since this disinterestedness must be common knowledge amongst the players, E and A have equally complex beliefs.  

We are interested in ME persuasion games in which the truth may emerge.  Is an ME persuasion game with a disinterested Jury sufficient to ensure such an outcome?  No.  there may be a fatal flaw in E's history that $A$ does not uncover and that the Jury does not see.  We have to suppose certain abilities on the part of $A$ and/or the Jury---namely, that if E has covered up some evidence or  falsely constructed evidence or has introduced an inconsistency in her history, that eventually A will uncover it.  Further, if there is an unexplained leap, an incoherence in the history, then $A$ will eventually find it.  Endowing $A$ with such capacities would suffice to ensure a history that is in E's winning condition to be the best possible approximation to the truth, a sort of Peircean ideal.  Even if we assume only that $A$ is a competent and skilled practitioner of her art, we have something like a good approximation of the truth for any history in E's winning condition.  We call a persuasion game with such a disinterested Jury and such a winning condition for $E$ an {\em ME truth game}.   

In an ME truth game, a player or a Jury may not be completely disinterested because of skewed priors.  But she may still be interested in finding out the truth and thus adjusting her priors in the face of evidence.  We put some constraints on the revision of beliefs of a {\em truth interested} player.  Suppose such a player $i$ has a prior $Pr^i$ on $a$\footnote{$a$ may be a fact, a history, a type etc. The exact nature of $a$ is not important for the argument.} such that $Pr^i(a) > 1/2$, but in a play $\play$ of an ME truth game it is revealed that $i$ has no confirming evidence for $a$ that the opponent $1-i$ cannot attack without convincing rebuttal.  Then a truth interested player $i$ should update her beliefs $Pr^i_\play$ after $\play$ so that $Pr^i_\play(a) = Pr^i(a|\play) \leq 1/2$.  On the other hand, if $i$ cannot rebut the confirming evidence that $1-i$ has for $\neg a$, then $Pr^i_\play(a) = Pr^i(a|\play) = 0$.  Where $\play$ is infinite, we put a condition on the prefixes $\play_n$ of $\play$: $Pr^i(a |\play) = \liminf_{n \rightarrow \infty} Pr^i(a|\play_n)$. 
Given our concepts of truth interested players and an ME truth game, we can show the following.
\begin{proposition} \label{agree} If the two players of a 2 history ME truth game $G$, have access to all the facts in ${\cal X}$, and are truth interested but have incompatible histories for ${\cal X}$ based on distinct priors, they will eventually agree to a common history for ${\cal X}$.
\end{proposition}
To prove this, we note that our players will note the disagreement and try to overcome it since they have a common interest, in the truth about ${\cal X}$.  Then it suffices to look at two cases: in case one, one player $i$ converges to the $1-i$'s beliefs in the ME game because $1-i$ successfully attacks the grounds on which $i$'s incompatible interpretation is based; in case two, neither $i$ nor $1-i$ is revealed to have good evidential grounds for their conflicting beliefs and so they converge to common revised beliefs that assign an equal probability to the prior beliefs that were in conflict.  Note that the difference with \cite{aumann:1976} is that we need to assume that players interested in the truth conditionalize upon outcomes of discussion in an ME game in the same way.  Players who do not do this need not ever agree.

There are interesting variants of an ME truth game where one has to do with approximations.  ME truth games are infinitary games, in which getting a winning history is something E may or may not achieve in the limit.  But typically we want the right, or ``good enough'' interpretation sooner rather than later. We can also appeal to discounted ME games developed in \cite{SEMDIAL}, in which the scores are assigned to individual discourse moves in context which diminish as the game progresses, to investigate cases where getting things right, or right enough, early on in an ME truth game is crucial.

In another variant of an ME truth game, which we call a {\em 2-history ME truth game}, we pit two biases one for E and one for A, and the two competing histories they engender, about a set of facts against each other.   Note that such a game is not necessarily win-lose as is the original ME truth game, because neither history the conversationalists develop and defend may satisfy the disinterested Jury.  That is, both E and A may lose in such a game.  Is it also possible that they both win?  Can both E and A revise their histories so that their opponents have in the end no telling attacks against their histories?  We think not at least in the case where the histories make or entail contradictory claims: in such a case they should both lose because they cannot defeat the opposing possibility.

Suppose $E$ wants to win an ME truth game and to construct a truthful history.  Let's assume that the set of facts ${\cal X}$ over which the history is constructed is finite.
What should she do?   Is it possible for her to win?  How hard is it for her to win?  Does $E$ have a winning strategy? As an ME truth game is win-lose, if the winning condition is Borel definable, it will be determined \cite{JPL}; either $E$ has a winning strategy or $A$ does.  Whether $E$ has a winning strategy or not is important: if she does, there is a method for finding an optimal history in the winning set; if she doesn't, an optimal history from the point of view of a truth-seeking goal in the ME truth game is not always attainable.   

To construct a history from ambiguous signals for a history over ${\cal X}$, the interpreter must rely on her beliefs about the situation and her interlocutors to estimate the right history.  So the question of getting at truthful interpretations of histories depends at least in part on the right answer to the question, 
what are the right beliefs about the situation and the participants that should be invoked in interpretation?  Given that beliefs are probabilistic, the space of possible beliefs is vast.  The right set of beliefs will typically form a very small set  with respect to the set of all possible beliefs about a typical conversational setting.  Assuming that one will be in such a position ``by default'' without any further argumentation is highly implausible, as a simple measure theoretic argument ensures that the set of possible interpretations are almost always biased away from a winning history in an ME truth game.  



What is needed for E-defensibility and a winning strategy in an ME truth game?  \cite{JPL} argued that consistency and coherence (roughly, the elements of the history have to be semantically connected in relevant ways \cite{asher:lascarides:2003}) are necessary conditions on all winning conditions and would thus apply to such histories.  A necessary additional property is completeness, an accounting of all or sufficiently many of the facts the history is claimed to cover.  We've also mentioned the care that has to be paid to the evidence and how it supports the history.  Finally, it became apparent when we considered a variant of an ME truth game in which two competing histories were pitted against each other that a winning condition for each player is that they must be able to defeat the opposing view or at least cast doubt on it.  

More particularly, truth seeking biases should provide predictive and explanatory power, which are difficult to define.  But we offer the following encoding of predictiveness and explanatory power as constraints on continuations of a given history in an ME truth game.  
\begin{definition} [Predictiveness]
A history $\hist$ developed in an ME game for a set of facts ${\cal X}$ is predictive just in case when $E$ is presented with a set of facts ${\cal Y}$ relevantly similar to ${\cal X}$, $\hist$ implies a E-defensible extension $\hist'$ of $\hist$ to all the facts in ${\cal Y}$.   
\end{definition}
A similar definition can be given for the explanatory power of a history.

Does $E$ have a strategy for constructing a truthful history that can guarantee all of these things?  Well, if the facts ${\cal X}$ it is supposed to relate are sufficiently simple or sufficiently unambiguous in the sense that they determine just one history and E is effectively able to build and defend such a history, then yes she does.   So  very simple cases like establishing whether your daughter has a snack for after school in the morning or not are easy to determine, and the history is equally simple, once you have the right evidence: yes she has a snack, or no she doesn't.    A text which is unambiguous similarly determines only one history, and linguistic competence should suffice to determine what that history is.  On the other hand, it is also possible that $E$ may determine the right history $\hist$ from a play $\play$ when $\hist$ depends on the type of the relevant players of $\play$.    For $E$ can have a true ``type'' for the players relevant to $\play$.  In general whether or not a player has a winning strategy will depend on the structure of the optimal history targeted, as well as on the resources and constraints on the players in an ME truth game.

In the more general case, however, whether $E$ has a winning strategy in an ME truth game become in general non trivial.   At least in a relative sort of way, E can construct a model satisfying her putative history at each stage to show consistency (relative to ZF or some other background theory); coherence can be verified by inspection over the finite discourse graph of the relevant history at each stage and ensuing attacks.  Finally completeness and evidential support can be guaranteed at each stage in the history's construction, if E has the right sort of beliefs.  If all this can be guaranteed at each stage, von Neumann's minimax theorem or its extension in \cite{blackwell:1956} guarantees that E has a winning strategy for E-defensibility.  


In future work, we plan to analyze in detail some complicated examples like the ongoing debate about climate, change where there is large scale scientific agreement but where disagreement exists because of distinct winning conditions.

 \section{Looking ahead} \label{sec:consequences}
 An ME truth game suggests a certain notion of truth: the truth is a winning history in an ME persuasion game with a disinterested Jury.  This is a Peircean ``best attainable'' approximation of the truth, an ''internal'' notion of truth based on consistency, coherence with the available evidence and explanatory and predictive power.   But we could investigate also a more external view of truth.   Such a view would suppose that the Jury has in its possession the ``true history over a set of facts ${\cal X}$, that the history eventually constructed by E should converge to within a certain margin of error in the limit.\footnote{To make this more precise, we need some measure of distance on histories \cite{morey:etal:2017,venant:etal:2013a,venant:etal:2013b}.  Then, assuming  a distance measure $d$, let $\hist^E_n$ and $\hist_n$ be the respective histories constructed over the prefix of a play of length $n$.  
 Letting $c \geq 0$ be a margin of error or cost, E's winning condition is:\\
 \ex. \label{win}
 $\lim_{n \rightarrow \omega} d(\hist_n, \hist^E_n) \leq c$
 
 }

We think ME games are a promising tool for investigating bias, and in this section we mention some possible applications and open questions that ME games might help us answer.  ME truth games allow us to analyze extant strategies for eliminating bias.   For instance, given two histories for a given set of facts, it is a common opinion that one finds a less biased history by splitting the difference between them.\footnote{For instance see, http://www.edu.gov.mb.ca/k12/cur/socstud/foundation\_gr9/blms/9-1-3g.pdf.}  This is a strategy perhaps distantly inspired by the idea that the truth lies in the golden mean between extremes.  But is this really true?   ME games should allow us to encode this strategy and find out.

Another connection that our approach can exploit is the one between games and reinforcement learning \cite{littman:1994,borgers:sarin:1997,hu:wellman:1998}.  While reinforcement learning is traditionally understood as a problem involving a single agent and is not powerful enough to understand the dynamics of competing biases of agents with different winning conditions, there is a direct connection made in \cite{borgers:sarin:1997} between evolutionary games with replicator dynamics and the stochastic learning theory of \cite{bush:mosteller:1955} with links to multiagent reinforcement learning.  \cite{littman:1994,hu:wellman:1998} provide a foundation for multiagent reinforcement learning in stochastic games.  The connection between ME games and stochastic and evolutionary games has not been explored but some victory conditions in ME games can be an objective that a replicator dynamics converges to, and epistemic ME games already encompass a stochastic component.  Thus, our research will be able to draw on relevant results in these areas.  


A typical assumption we make as scientists is that rationality would lead us to always prefer to have a more complete and more accurate history for our world.   But bias isn't so simple, as an analysis of ME games can show.   ME games are played for many purposes with non truth-seeking biases that lead to histories that are not a best approximation to the truth may be the rational or optimal choice, if the winning condition in the game is other than that defined in an ME truth game.  This has real political and social relevance; for example, a plausible hypothesis is that those who argue that climate change is a hoax are building an alternative history, not to get at the truth but for other political purposes.

Even being a truth interested player can at least initially fail to generate histories that are in the winning condition of an ME truth game.  Suppose E, motivated by truth interest, has constructed for facts ${\cal X}$ a history $\hist$  that meets constraints including coherence, consistency, and completeness, and it provides explanatory and predictive power for at least a large subset ${\cal Z}$ of ${\cal X}$.  E's conceptualization of ${\cal X}$ can still go wrong, and E may fail to have a winning strategy in interesting ways.  First, $\hist$ can mischaracterize ${\cal X}$ with high confidence in virtue of evidence only from ${\cal Z}$ \cite{lakkaraju:etal:2016};\footnote{This option encapsulates the problem of optimizing the decision to exploit a bias that has a certain ``local'' optimality or to explore the space of possible biases further.  There is a large literature on this issue \cite{whittle:1980,lai:robbins:1985,banks:sundaram:1994,burnetas:katehakis:1997,auer:etal:2002,garivier:cappe:2011}.}   Especially if ${\cal Z}$ is large and hence $\hist$ is just simply very ``long'',  it is intuitively more difficult even for truth seeking players to come to accept that an alternative history is the correct one.  Second, $\hist$ may lack or be incompatible with concepts that would be needed to be aware of facts in ${\cal X} \setminus {\cal Z}$.      \cite{asher:paul:2013,venant:2016} investigate a special case of this, a case of unawareness.  To succeed E would have to learn the requisite concepts first.  


All of this has important implications for learning.  We can represent learning as the following ME games. It is common to represent making a prediction Y from data X as a zero sum game between our player E and Nature: E wins if for data X provided by Nature, E makes a prediction that the Jury judges to be correct.  More generally, an iterated learning process is a repeated zero sum game, in which E makes predictions in virtue of some history, which one might also call a model or a set of hypotheses; if she makes a correct prediction at round n, she reinforces her beliefs in her current history; if she makes a wrong prediction, she adjusts it. The winning condition may be defined in terms of some function of the scores at each learning round or in terms of some global convergence property. Learning conceived in this way is a variant of a simple ME truth game in which costs are assigned to individual discourse moves as in discounted ME games.

In an ME truth game, where E develops a history $\hist^*$ over a set of facts $\cal{X}$ while A argues for an alternative history $\hist$ over ${\cal X}$, 
A can successfully defend history $\hist$ as long as either the true history $\hist^*$ is (a) not learnable or (b) not uniquely learnable.  In case (a), E cannot convince the Jury that $\hist^*$ is the right history; in case (b) A can justify $\hist$ as an alternative interpretation.  
Consider the bias of a hardened climate change skeptic:  the ME model predicts that simply presenting new facts to the agent will not induce him to change his history, even if to a disinterested Jury his history is clearly not in his winning condition.  He may either simply refuse to be convinced because he is not truth interested, or because he thinks his alternative history $\hist$ can explain all of the data in ${\cal X}$ just as well as E's climate science history $\hist^*$.  
Thus, ME games open up an unexplored research area of {\em unlearnable} histories for certain agents.

\section{Conclusions}\label{sec:conclusions}

In this paper, we have put forward the foundations of a formal model of interpretive bias.  
Our approach differs from philosophical and AI work on dialogue that links dialogue understanding to the recovery of speaker intentions and beliefs \cite{grice:1975,grice:1969}.  Studies of multimodal interactions in Human Robot Interaction (HRI) have also followed the Gricean tradition \cite{perzanowski:etal:2001,chambers:etal:2005,foster:petrick:2014}.  \cite{asher:lascarides:2013,JPL,lepore:stone:2015}), offer many reasons why a Gricean program for dialogue understanding is difficult for dialogues in which there is not a shared task and a strong notion of co-operativity.  Our model is not in the business of intention and belief recovery, but rather works from what contents agents explicitly commit to with their actions, linguistic and otherwise, to determine a rational reconstruction of an underlying interpretive bias and what goals a bias would satisfy.   In this we also go beyond what current theories of discourse structure like SDRT can accomplish.   

Our theoretical work also requires an empirical component on exactly how bias is manifested to be complete.  This has links to the recent interest in fake news.\footnote{See \cite{lee:solomon:1990} and more recent work by {\em Google News}, {\em FirstDraft News}, and  {\em Cross Check} in France, as well as Hoaxy (https://hoaxy.iuni.iu.edu/).} 
Modeling interpretive bias can help in detecting fake news by providing relevant types to check in interpretation and by providing an epistemic foundation for fake news detection by exploiting ME truth games where one can draw from various sources to check the credibility of a story.   In a future paper, we intend to investigate these connections thoroughly.   


{

\bibliographystyle{splncs04}
\bibliography{erc2}

\begin{thebibliography}{10}
\providecommand{\url}[1]{\texttt{#1}}
\providecommand{\urlprefix}{URL }
\providecommand{\doi}[1]{https://doi.org/#1}

\bibitem{asher:lascarides:2013}
Asher, N., Lascarides, A.: Strategic conversation. Semantics and Pragmatics
  \textbf{6}(2),  http:// dx.doi.org/10.3765/sp.6.2. (2013)

\bibitem{SEMDIAL}
Asher, N., Paul, S.: Evaluating conversational success: Weighted message
  exchange games. In: Hunter, J., Simons, M., Stone, M. (eds.) 20th workshop on
  the semantics and pragmatics of dialogue (SEMDIAL). New Jersey, USA (July
  2016)

\bibitem{asher:1993}
Asher, N.: Reference to Abstract Objects in Discourse. Kluwer Academic
  Publishers (1993)

\bibitem{asher:lascarides:2003}
Asher, N., Lascarides, A.: Logics of Conversation. Cambridge University Press
  (2003)

\bibitem{asher:paul:2013}
Asher, N., Paul, S.: Conversations and incomplete knowledge. In: Proceedings of
  Semdial Conference. pp. 173--176. Amsterdam (December 2013)

\bibitem{IACL}
Asher, N., Paul, S.: Conversation and games. In: Ghosh, S., Prasad, S. (eds.)
  Logic and Its Applications: 7th Indian Conference, ICLA 2017, Kanpur, India,
  January 5-7, 2017, Proceedings. vol. 10119, pp. 1--18. Springer, Kanpur,
  India (January 2017)

\bibitem{JOLLI}
Asher, N., Paul, S.: Strategic conversation under imperfect information:
  epistemic {M}essage {E}xchange games (2017), accepted for publication in {\em
  Journal of Logic, Language and Information}

\bibitem{JPL}
Asher, N., Paul, S., Venant, A.: Message exchange games in strategic
  conversations. Journal of Philosophical Logic  \textbf{46.4},  355--404
  (2017), \url{http://dx.doi.org/10.1007/s10992-016-9402-1}

\bibitem{auer:etal:2002}
Auer, P., Cesa-Bianchi, N., Fischer, P.: Finite-time analysis of the multiarmed
  bandit problem. Machine learning  \textbf{47}(2-3),  235--256 (2002)

\bibitem{aumann:1976}
Aumann, R.J.: Agreeing to disagree. The Annals of Statistics  \textbf{4}(6),
  1236--1239 (1976)

\bibitem{banks:sundaram:1994}
Banks, J.S., Sundaram, R.K.: Switching costs and the gittins index.
  Econometrica: Journal of the Econometric Society pp. 687--694 (1994)

\bibitem{baron:2000}
Baron, J.: Thinking and deciding. Cambridge University Press (2000)

\bibitem{battigalli:2003}
Battigalli, P.: Rationalizability in infinite, dynamic games with incomplete
  information. Research in Economics  \textbf{57}(1),  1--38 (2003)

\bibitem{berger:etal:1996}
Berger, A.L., Pietra, V.J.D., Pietra, S.A.D.: A maximum entropy approach to
  natural language processing. Computational linguistics  \textbf{22}(1),
  39--71 (1996)

\bibitem{besnard:hunter:2008}
Besnard, P., Hunter, A.: Elements of argumentation, vol.~47. MIT press
  Cambridge (2008)

\bibitem{blackwell:1956}
Blackwell, D.: An analog of the minimax theorem for vector payoffs. Pacific
  Journal of Mathematics  \textbf{6}(1), ~1--8 (1956)

\bibitem{borgers:sarin:1997}
B{\"o}rgers, T., Sarin, R.: Learning through reinforcement and replicator
  dynamics. Journal of Economic Theory  \textbf{77}(1),  1--14 (1997)

\bibitem{burnetas:katehakis:1997}
Burnetas, A.N., Katehakis, M.N.: Optimal adaptive policies for markov decision
  processes. Mathematics of Operations Research  \textbf{22}(1),  222--255
  (1997)

\bibitem{burnett:2017}
Burnett, H.: Sociolinguistic interaction and identity construction: The view
  from game-theoretic pragmatics. Journal of Sociolinguistics  \textbf{21}(2),
  238--271 (2017)

\bibitem{bush:mosteller:1955}
Bush, R.R., Mosteller, F.: Stochastic models for learning. John Wiley \& Sons,
  Inc. (1955)

\bibitem{cadilhac:etal:2011}
Cadilhac, A., Asher, N., Benamara, F., Lascarides, A.: Commitments to
  preferences in dialogue. In: Proceedings of the 12th Annual SIGDIAL Meeting
  on Discourse and Dialogue. pp. 204--215 (2011)

\bibitem{cadilhac:etal:2013}
Cadilhac, A., Asher, N., Benamara, F., Lascarides, A.: Grounding strategic
  conversation: Using negotiation dialogues to predict trades in a win-lose
  game. In: Proceedings of EMNLP. pp. 357--368. Seattle (2013)

\bibitem{cadilhac:etal:2012a}
Cadilhac, A., Asher, N., Benamara, F., Popescu, V., Seck, M.: Preference
  extraction form negotiation dialogues. In: Biennial European Conference on
  Artificial Intelligence (ECAI) (2012)

\bibitem{chambers:etal:2005}
Chambers, N., Allen, J., Galescu, L., Jung, H.: A dialogue-based approach to
  multi-robot team control. In: The 3rd International Multi-Robot Systems
  Workshop. Washington, DC (2005)

\bibitem{dung:1995}
Dung, P.M.: On the acceptability of arguments and its fundamental role in
  nonmonotonic reasoning, logic programming and n-person games. Artificial
  intelligence  \textbf{77}(2),  321--357 (1995)

\bibitem{erev:etal:1994}
Erev, I., Wallsten, T.S., Budescu, D.V.: Simultaneous over-and underconfidence:
  The role of error in judgment processes. Psychological review
  \textbf{101}(3), ~519 (1994)

\bibitem{foster:petrick:2014}
Foster, M.E., Petrick, R.P.A.: Planning for social interaction with sensor
  uncertainty. In: The {ICAPS} 2014 Scheduling and Planning Applications
  Workshop ({SPARK}). pp. 19--20. Portsmouth, New Hampshire, USA (Jun 2014)

\bibitem{garivier:cappe:2011}
Garivier, A., Capp{\'e}, O.: The kl-ucb algorithm for bounded stochastic
  bandits and beyond. In: COLT. pp. 359--376 (2011)

\bibitem{glazer:rubinstein:2004}
Glazer, J., Rubinstein, A.: On optimal rules of persuasion. Econometrica
  \textbf{72}(6),  119--123 (2004)

\bibitem{grice:1969}
Grice, H.P.: Utterer's meaning and intentions. Philosophical Review
  \textbf{68}(2),  147--177 (1969)

\bibitem{grice:1975}
Grice, H.P.: Logic and conversation. In: Cole, P., Morgan, J.L. (eds.) Syntax
  and Semantics Volume 3: Speech Acts, pp. 41--58. Academic Press (1975)

\bibitem{grosz:sidner:1986}
Grosz, B., Sidner, C.: Attention, intentions and the structure of discourse.
  Computational Linguistics  \textbf{12},  175--204 (1986)

\bibitem{harsanyi:1967}
Harsanyi, J.C.: Games with incomplete information played by “bayesian”
  players, parts i-iii. Management science  \textbf{14},  159--182 (1967)

\bibitem{henderson:mccready:2017}
Henderson, R., McCready, E.: Dogwhistles and the at-issue/non-at-issue
  distinction. Published on Semantics Archive  (2017)

\bibitem{hilbert:2012}
Hilbert, M.: Toward a synthesis of cognitive biases: how noisy information
  processing can bias human decision making. Psychological bulletin
  \textbf{138}(2), ~211 (2012)

\bibitem{hintzman:1984}
Hintzman, D.L.: Minerva 2: A simulation model of human memory. Behavior
  Research Methods, Instruments, \& Computers  \textbf{16}(2),  96--101 (1984)

\bibitem{hintzman:1988}
Hintzman, D.L.: Judgments of frequency and recognition memory in a
  multiple-trace memory model. Psychological review  \textbf{95}(4), ~528
  (1988)

\bibitem{hu:wellman:1998}
Hu, J., Wellman, M.P.: Multiagent reinforcement learning: theoretical framework
  and an algorithm. In: ICML. vol.~98, pp. 242--250 (1998)

\bibitem{hunter:etal:2017}
Hunter, J., Asher, N., Lascarides, A.: Situated conversation (2017), submitted
  to {\em Semantics and Pragmatics}

\bibitem{khoo:2017}
Khoo, J.: Code words in political discourse. Philosophical Topics
  \textbf{45}(2),  33--64 (2017)

\bibitem{konek:2016}
Konek, J.: Probabilistic knowledge and cognitive ability. Philosophical Review
  \textbf{125}(4),  509--587 (2016)

\bibitem{lai:robbins:1985}
Lai, T.L., Robbins, H.: Asymptotically efficient adaptive allocation rules.
  Advances in applied mathematics  \textbf{6}(1),  4--22 (1985)

\bibitem{lakkaraju:etal:2016}
Lakkaraju, H., Kamar, E., Caruana, R., Horvitz, E.: Discovering blind spots of
  predictive models: Representations and policies for guided exploration. arXiv
  preprint arXiv:1610.09064  (2016)

\bibitem{lee:solomon:1990}
Lee, M., Solomon, N.: Unreliable Sources: A Guide to Detecting Bias in News
  Media. Lyle Smart, New York (1990)

\bibitem{lepore:stone:2015}
Lepore, E., Stone, M.: Imagination and Convention: Distinguishing Grammar and
  Inference in Language. Oxford University Press (2015)

\bibitem{littman:1994}
Littman, M.L.: Markov games as a framework for multi-agent reinforcement
  learning. In: Proceedings of the eleventh international conference on machine
  learning. vol.~157, pp. 157--163 (1994)

\bibitem{morey:etal:2017}
Morey, M., Muller, P., Asher, N.: A dependency perspective on rst discourse
  parsing and evaluation (2017), submitted to {\em Computational Linguistics}

\bibitem{moss:2013}
Moss, S.: Epistemology formalized. Philosophical Review  \textbf{122}(1),
  1--43 (2013)

\bibitem{perret:etal:2016}
Perret, J., Afantenos, S., Asher, N., Morey, M.: Integer linear programming for
  discourse parsing. In: Proceedings of the 2016 Conference of the North
  American Chapter of the Association for Computational Linguistics: Human
  Language Technologies. pp. 99--109. Association for Computational
  Linguistics, San Diego, California (June 2016),
  \url{http://www.aclweb.org/anthology/N16-1013}

\bibitem{perzanowski:etal:2001}
Perzanowski, D., Schultz, A., Adams, W., Marsh, E., Bugajska, M.: Building a
  multimodal human-robot interface. Intelligent Systems  \textbf{16}(1),
  16--21 (2001)

\bibitem{potts:2005}
Potts, C.: The logic of conventional implicatures. Oxford University Press
  Oxford (2005)

\bibitem{recanati:2004}
Recanati, F.: Literal Meaning. Cambridge University Press (2004)

\bibitem{sperber:wilson:1986}
Sperber, D., Wilson, D.: Relevance. Blackwells (1986)

\bibitem{stanley:2015}
Stanley, J.: How propaganda works. Princeton University Press (2015)

\bibitem{tversky:kahneman:1973}
Tversky, A., Kahneman, D.: Availability: A heuristic for judging frequency and
  probability. Cognitive psychology  \textbf{5}(2),  207--232 (1973)

\bibitem{tversky:kahneman:1975}
Tversky, A., Kahneman, D.: Judgment under uncertainty: Heuristics and biases.
  In: Utility, probability, and human decision making, pp. 141--162. Springer
  (1975)

\bibitem{tversky:kahneman:1983}
Tversky, A., Kahneman, D.: Extensional versus intuitive reasoning: The
  conjunction fallacy in probability judgment. Psychological review
  \textbf{90}(4), ~293 (1983)

\bibitem{tversky:kahneman:1985}
Tversky, A., Kahneman, D.: The framing of decisions and the psychology of
  choice. In: Environmental Impact Assessment, Technology Assessment, and Risk
  Analysis, pp. 107--129. Springer (1985)

\bibitem{venant:2016}
Venant, A.: Structures, Semantics and Games in Strategic Conversations. Ph.D.
  thesis, Universit\'e Paul Sabatier, Toulouse (2016)

\bibitem{venant:etal:2013b}
Venant, A., Asher, N., Muller, P., Denis, P., Afantenos, S.: Expressivity and
  comparison of models of discourse structure. In: Proceedings of the SIGDIAL
  2013 Conference. pp. 2--11. Association for Computational Linguistics, Metz,
  France (August 2013), \url{http://www.aclweb.org/anthology/W13-4002}

\bibitem{venant:etal:2013a}
Venant, A., Degremont, C., Asher, N.: Semantic similarity. In: LENLS 10. Tokyo,
  Japan (2013)

\bibitem{walton:1984}
Walton, D.N.: Logical dialogue-games. University Press of America (1984)

\bibitem{whittle:1980}
Whittle, P.: Multi-armed bandits and the gittins index. Journal of the Royal
  Statistical Society. Series B (Methodological) pp. 143--149 (1980)

\bibitem{wilkinson:klaes:2012}
Wilkinson, N., Klaes, M.: An introduction to behavioral economics. Palgrave
  Macmillan (2012)

\end{thebibliography}

\end{document}